\documentclass[preprint, 12pt]{elsart}
\usepackage{amsmath,amsfonts}
\usepackage{algorithmic}
\usepackage{algorithm}
\usepackage{array}
\usepackage{textcomp}
\usepackage{stfloats}
\usepackage{url}
\usepackage{ifthen}
\usepackage{verbatim}
\usepackage{graphicx}
\usepackage{booktabs}
\usepackage{makecell}
\usepackage{multirow}
\usepackage{xcolor}
\definecolor{my_purple}{HTML}{92268F}
\definecolor{matany}{HTML}{98CC70}
\usepackage{hyperref}
\usepackage{wrapfig}
\usepackage{subfig}
\newcommand{\thename}{MatAny}
\newcommand{\themam}{Matte Anything model}
\newcommand{\eq}[1]{$#1$}
\usepackage{colortbl}

\begin{document}
\begin{sloppypar}
\begin{frontmatter}
\title{Matte Anything: Interactive Natural Image Matting with Segment Anything Model}
\author[hustei]{Jingfeng Yao}
\author[hustei]{Xinggang Wang\corref{corr}}\ead{xgwang@hust.edu.cn}
\author[hustei]{Lang Ye}
\author[hustei]{Wenyu Liu}

\cortext[corr]{Corresponding author}
\address[hustei]{School of Electronic Information and Communications, Huazhong University of Science and Technology, Wuhan 430074, China}

\begin{abstract}
Natural image matting algorithms aim to predict the transparency map (alpha-matte) with the trimap guidance. However, the production of trimap often requires significant labor, which limits the widespread application of matting algorithms on a large scale. To address the issue, we propose \themam{} (\thename{}), an interactive natural image matting model that could produce high-quality alpha-matte with various simple hints. The key insight of \thename{} is to generate pseudo trimap automatically with contour and transparency prediction. In our work, we leverage vision foundation models to enhance the performance of natural image matting. Specifically, we use the segment anything model to predict high-quality contour with user interaction and an open-vocabulary detector to predict the transparency of any object. Subsequently, a pre-trained image matting model generates alpha mattes with pseudo trimaps. \thename{} is the interactive matting algorithm with the most supported interaction methods and the best performance to date. It consists of orthogonal vision models without any additional training. We evaluate the performance of \thename{} against several current image matting algorithms. \thename{} has 58.3\% improvement on MSE and 40.6\% improvement on SAD compared to the previous image matting methods with simple guidance, achieving new state-of-the-art (SOTA) performance. The source codes and pre-trained models are available at \url{https://github.com/hustvl/Matte-Anything}.


\end{abstract}
\end{frontmatter}

\section{Introduction}
\label{sec:intro}

Natural image matting is a prominent computer vision task with significant implications~\cite{knnmatting2013, closed-form, matting_survey, matting2013, hematting2011}. Its primary objective is to accurately predict the transparency map, commonly referred to as the alpha matte, for an object in a given image. Unlike image segmentation~\cite{maskformer, mask2former, oneformer}, natural image matting offers more precise predictions and excels in handling transparent objects, such as glasses. Consequently, it finds widespread applications in the generation of posters and the creation of visual effects for movies. Recently, image matting algorithms~\cite{DIM, indexnet, GCAMatting, MGM, fba, matteformer, vitmatte} with deep learning have achieved remarkable performance.

\begin{figure}[tbp]
    \centering
    \includegraphics[width=\textwidth]{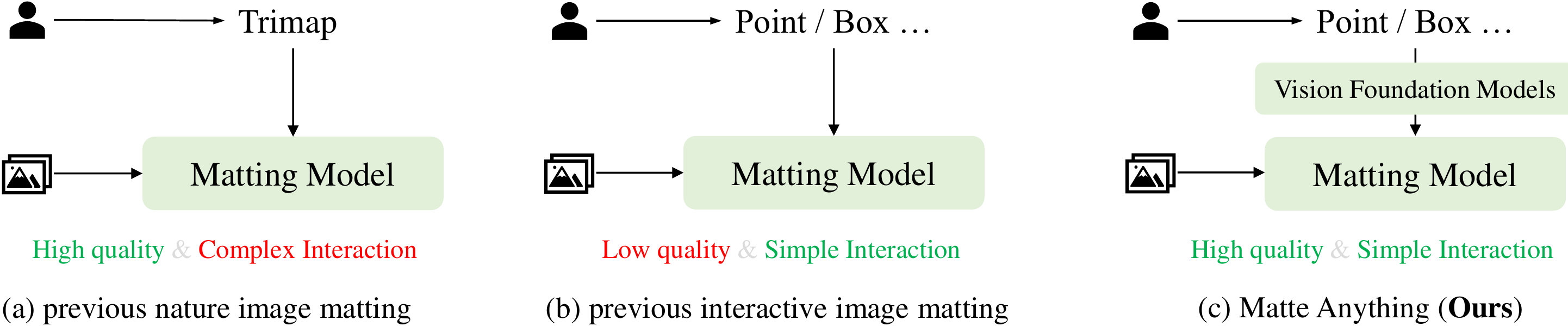}
    \caption{\textbf{Comparison between Matte Anything and previous methods.} Matte Anything utilizes vision foundational models, such as Segment Anything Models~\cite{SAM}, Open Vocabulary Detection Models~\cite{groundingdino}, etc., achieving interactively simple and high-quality natural image matting.}
    \label{fig:first}
\end{figure}

\begin{figure*}
    \centering
    \includegraphics[width=1\linewidth]{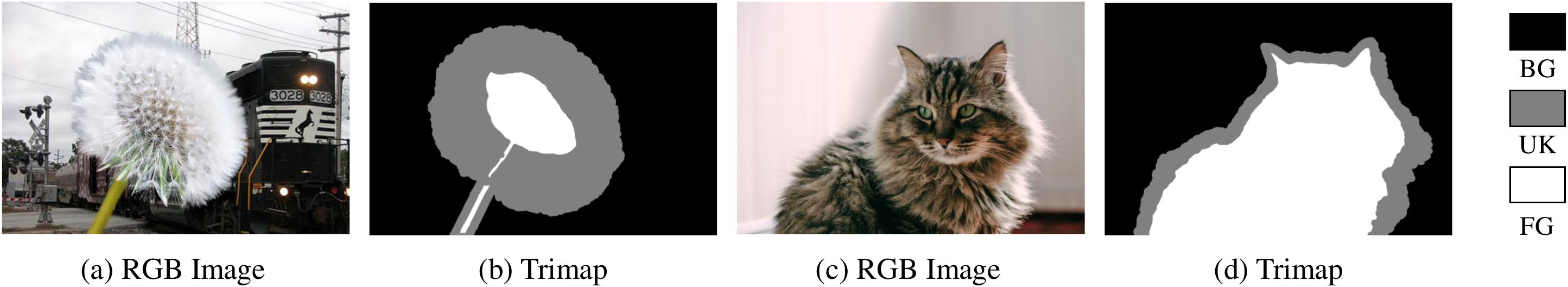}
    \caption{\textbf{RGB Image and Trimap.} The trimap divides an image into foreground (FG), background (BG), and an unknown region (UK). Algorithms based on the trimap only need to predict the unknown region. This reduces the algorithmic complexity but increases the interactive cost.}
    \label{fig:trimap}
\end{figure*}

These image matting algorithms, illustrated as Figure~\ref{fig:first}~(a), aim to predict the transparency of an object by utilizing the guidance provided by a trimap. As shown in Figure~\ref{fig:trimap}, the trimap is a hint map that is manually labeled for the purpose of image matting. It effectively divides an image into three distinct regions: foreground, background, and an unknown region. In the current state-of-the-art natural image matting methods~\cite{vitmatte}, both the original image and its corresponding trimap are simultaneously used as input. However, despite the excellent matting results achieved by these methods, they have not yet become mainstream approaches widely adopted on a large scale. One crucial limitation of these methods is the high labor cost associated with generating trimaps.

An intuitive solution is to perform image matting with minimal or even no interaction, using straightforward methods, as illustrated in Figure~\ref{fig:first}~(b). One approach ~\cite{user_click, unimatting, DIIM} is to incorporate additional inexpensive cues, such as bounding boxes, to guide image matting. These methods often utilize the same training dataset~\cite{DIM} as trimap-based methods. The problem with such a scheme is that after losing the information-rich trimap, a single network is required to simultaneously tackle the challenging tasks of image segmentation, transparency assessment, and image matting using a single dataset. This severely restricts the overall performance and generalization capabilities of the network. Another approach~\cite{HAttMatting, LFM} involves performing image matting using only RGB images without any guidance. However, a significant drawback of these methods is that they treat the result of image matting as an image-level task. A challenge would arise when faced with multiple instances, as it becomes difficult for them to determine which object should be matted. Hence, the problem of high-performance matting with simple interaction remains to be solved.

Recently, vision foundation models have been driving rapid advancements in a multitude of visual tasks. For instance, the Segment Anything Model (SAM)~\cite{SAM} is an interactive segmentation model that can generate corresponding segmentations based on user-provided points or bounding boxes. Some works have applied SAM to various domains such as object tracking~\cite{trackanything}, image generation~\cite{inpaint_anything}, medical segmentation~\cite{MedSAM}, demonstrating its immense value. Another example is the Open Vocabulary Detection (OVD) model, Grounding DINO~\cite{groundingdino}, which combines the CLIP~\cite{clip} model and the detection model DINO~\cite{dino} to generate bounding boxes for corresponding objects based on user's textual prompts. This makes it possible to actively obtain the positional information of arbitrary objects.

Back to natural image matting, the essence of it lies in simultaneously capturing the segmentation, transparency, and detailed information from the image. The key to the high performance of the trimap-based methods is that the trimap includes segmentation and transparency information directly provided by the user. So, these motivate us to think: \textit{Can we use these powerful vision foundation models instead of manually drawn trimaps to achieve high-performance, simply interactive image matting?}

To address the aforementioned challenges, we propose \themam{} (\thename{}), a high-performance image matting framework with simple interaction by leveraging vision foundation models (as shown in Figure~\ref{fig:first}~(c)). The key idea of our approach is to use the segmentation and transparency information generated by the vision foundation model to generate a pseudo-trimap. Accordingly, we approach image matting as a downstream task of image segmentation and transparent object detection. In \thename{}, we leverage task-specific vision models to enhance the performance of natural image matting. Firstly, we employ the Segment Anything Model (SAM)~\cite{SAM} to generate a high-quality mask for the target instance. Subsequently, we utilize the open-vocabulary object detection model, i.e., GroudingDINO~\cite{groundingdino} to detect commonly occurring transparent objects. Pseudo-trimaps are then generated based on the segmentation and transparent object detection results, which are subsequently inputted into natural image matting models, e.g., ViTMatte~\cite{vitmatte}. The ViTMatte model is the state-of-the-art class-agnostic matting method, is implemented by efficiently adapting pre-trained ViTs, and has strong generalization ability. The three parts, SAM, GroundingDINO, and ViTMatte, are decoupled and require no additional training.

Leveraging the strong capability of these task-specific models, \thename{} yields unprecedented matting capabilities. It has superior properties. First, it can support the largest variety of interactive lists, including points, boxes, scribbles and texts. Second, it can be further refined by very simple means (mouse clicks). Meanwhile, it shows strong generalization and zero-shot capabilities on real and specific matting scenarios. We evaluate \thename{} on four image matting datasets and the results demonstrate its great potential. On the most widely-used matting benchmark Composition-1k~\cite{DIM}, our method outperforms other \textit{x}-guide matting methods (\textit{x} could be points, boxes, and so on). Compared with previous SOTA methods~\cite{unimatting}, \thename{} achieves 58.3\% improvement on the MSE and 40.6\% improvement on SAD. \thename{} also achieves competitive results to trimap-based matting methods with simple refinement. Meanwhile, on real-world images, we find that \thename{} exhibits strong generalization performance and zero-shot performance. Even without any fine-tuning, it can still achieve very competitive performance. 

To sum up, our contributions could be summarized as follows:
\begin{itemize}
    \item We present Matte Anything (\thename{}), a high-performance and simply interactive matting framework composed of decoupled vision models that require no additional training. To the best of our knowledge, \thename{} is the first image matting method leveraging the great powers of vision foundation models, such as SAM.
    \item We propose a trimap generation strategy based on vision foundation models. This strategy enables the generation of high-quality and adaptive pseudo trimaps with minimal user input, significantly reducing the manual effort required for trimap production.
    \item We evaluate the performance and generalization ability of Matte Anything using 4 image matting benchmarks. 
    \thename{} achieves SOTA performance (58.3\% improvement on the MSE) compared to previous x-guided matting methods and very comparable performance with trimap-guided methods. 
    The results demonstrate its significant potential in achieving high-quality matting results across diverse datasets.
\end{itemize}

\section{Related Work}

\subsection{Natural Image Matting} 

\noindent
\textbf{Trimap-based Natural Image Matting.}
Natural image matting is class-agnostic, as it aims to segment any selected instances within an image. Deep learning algorithms~\cite{DIM, indexnet, GCAMatting, fba, MGM, matteformer, vitmatte}, represented by DIM~\cite{DIM} (Deep Image Matting), have achieved significant advancements in image matting. 
Subsequently, numerous outstanding matting methods have emerged. 
Context-aware Matting~\cite{CAM} addresses issues related to inadequate semantic information extraction and fusion by proposing a dual network structure that combines global semantics and local features.
IndexNet~\cite{indexnet} introduces an index-guided upsampling strategy to tackle matting problems.
MGMatting~\cite{MGM} further enhances the detail representation capability of matting models by using a progressively refining decoder.
MatteFormer~\cite{vitmatte} improves the network's perception of trimap information by designing prior tokens.
ViTMatte~\cite{vitmatte} achieves the current best matting performance by adapting a pre-trained ViT model to the matting task.
However, these algorithms have a common requirement for trimap as guidance information, which restricts their applicability to a wide range of scenarios. 

\noindent
\textbf{Trimap-free Natural Image Matting.}
Some methods~\cite{user_click, unimatting, DIIM} attempt to perform image matting using simplistic guidance, such as points or bounding boxes, and there are even attempts~\cite{LFM, HAttMatting} to achieve matting without any explicit guidance. However, these methods often experience notable performance degradation and poor generalization capabilities, primarily due to the limited datasets available for training and the restricted number of parameters employed. There is an urgent need for an efficient and high-performance natural image matting algorithm with just simple guidance.

\vspace{-3mm}
\subsection{Foundation Models}

\noindent
\textbf{Segment Anything.}
Large Language Models (LLMs)~\cite{GPT3, gpt-4, llama} have garnered significant attention in both the Natural Language Processing (NLP) and Computer Vision (CV) fields, as exemplified by models like GPT-4~\cite{gpt-4} and LLaMA~\cite{llama}. Researchers have observed the immense power of scaling up deep learning models, where foundational models trained on extensive data can unlock boundless possibilities for downstream tasks.
Recently, Kirillov et al. have introduced the Segment Anything Model (SAM)~\cite{SAM} as a segmentation foundation model in computer vision, capable of segmenting any object based on user prompts. Researchers have successfully leveraged SAM to enhance various downstream tasks, such as image inpainting~\cite{inpaint_anything}, image generation~\cite{controlnet, stablediffusion}, and so on \cite{SAMvsBET,samm,clipsurgery,inpaint,anything3d,trackanything,cen2023segment,MedSAM,SAM-Track}. However, to the best of our knowledge, there has been no prior investigation into the potential of SAM in the context of image matting tasks.

\noindent
\textbf{Open Vocabulary Detection.}
Open Vocabulary (OV) detection~\cite{OV-DETR,ViLD,GLIP,DetCLIP} focuses on identifying the bounding box of diverse categories without predefined constraints. Leveraging the foundation model CLIP~\cite{clip}, researchers have achieved notable zero-shot performance on detection datasets~\cite{mscoco, lvis, odinw}, showcasing its robust generalization capability. Furthermore, recent advancements in the field, such as GroundingDINO~\cite{groundingdino}, have demonstrated impressive results, surpassing 50 mAP on the COCO~\cite{mscoco} dataset through the fusion of vision and language modalities at multiple stages. These innovations have found broad applications in image generation, image editing, and image segmentation. Building upon these achievements, our work in Matte Anything introduces a novel perspective on OV detection for image matting tasks.

\section{Methods}
\label{sec:method}

\begin{figure*}[t]
    \centering
    \includegraphics[width=1.0\textwidth]{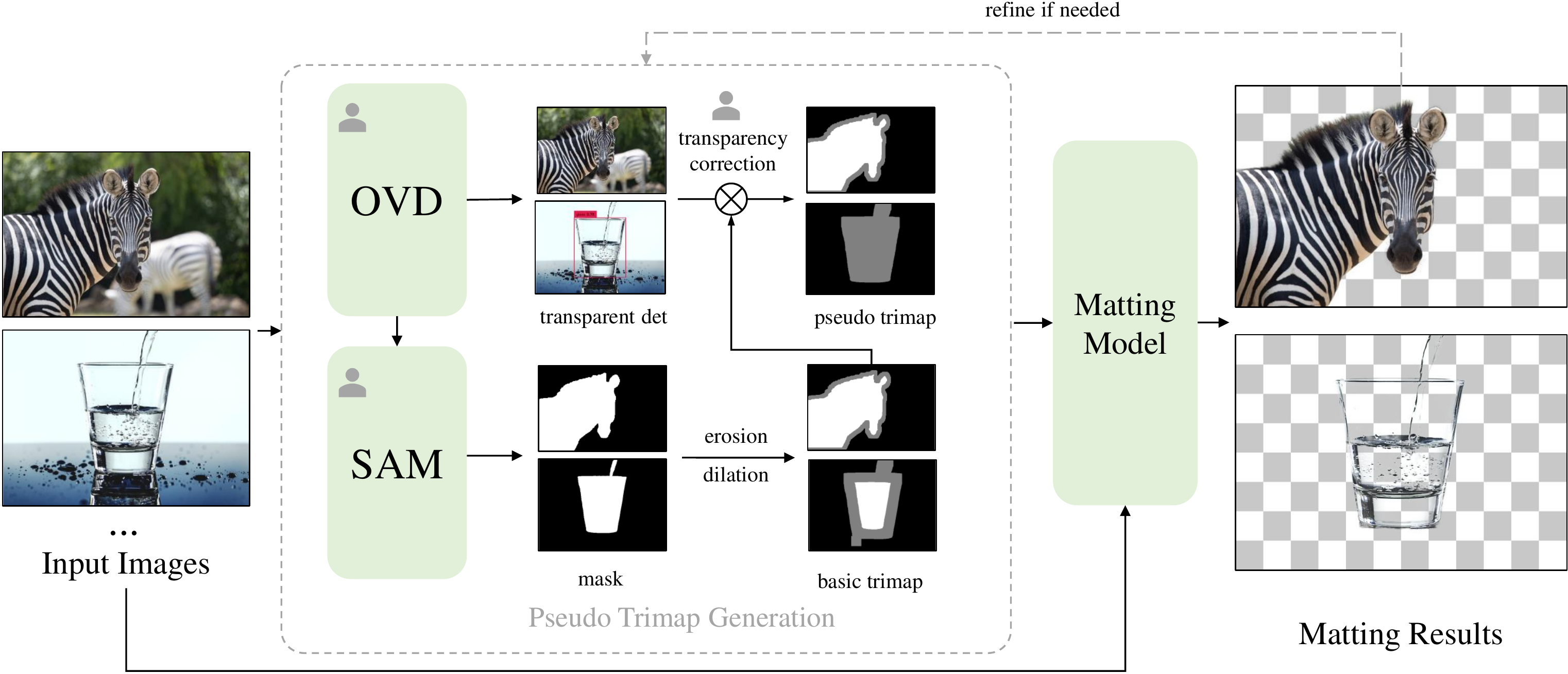}
    \caption{Overall architecture of \themam{} (\thename{}). Its core idea is to interactively utilize visual backbone models to generate a pseudo-trimap, which is then fed into a pre-trained trimap-based model to produce high-quality matting results. ``\textit{OVD}" denotes Open-Vocabulary Detectors, ``\textit{SAM}" denotes Segment Anything Models.}
    \label{fig:overall}
\end{figure*}

We propose Matte Anything, a high-performance and simply interactive natural image matting model. The key insight of our method is to generate pseudo-trimap with vision foundation models~\cite{SAM, groundingdino}. In this section, we introduce each part our proposed methods and the pseudo-trimap generation strategy. 

\subsection{Preliminaries: Priors of Trimap}
\label{sec:prior_trimap}
For ease of comprehension, we first give a brief introduction to trimap in image matting. As discussed above, trimap is a hint map to provide information on the foreground, background, and unknown region for a given image. There are two main priors of an accurate trimap. 

\emph{\textbf{Prior 1}, an image matting algorithm will predict only in unknown regions of trimap}.  That means a predicted alpha matte will be like Equation~\ref{eq:trimap}:
\begin{equation}
    \alpha(x, y)=
    \begin{cases}
        1 & if~(x, y) \in F \\
        M(i, t) & if~(x, y) \in U \\
        0 & if~(x, y) \in B
    \end{cases}
    \label{eq:trimap}
\end{equation}
where $F$, $U$, and $B$ denote the regions of foreground, unknown, and background in trimap respectively. $M$ denotes the image matting model, $i, t$ denotes input image and trimap. 

\emph{\textbf{Prior 2}, a transparent region will not be the foreground region in trimap}. Since the pixel's alpha value in the transparent region should be in $[0, 1)$. If a transparent region is labeled as foreground in trimap, it will conflict with Equation~\ref{eq:trimap}.

Based on these two main priors, we start to introduce our \themam{}.

\subsection{Overall Architecture}

As depicted in Figure~\ref{fig:overall}, \thename{} comprises three key components: a natural image matting model, a Segment Anything Model (SAM)~\cite{SAM}, and an Open Vocabulary detector (OVD)~\cite{groundingdino}. It's important to note that these models are not specific to certain ones but represent a category of models. Therefore, the performance of Matte Anything can improve as the performance of vision foundation models improves.

We use the OV detector to generate bounding boxes with given text for SAM and detect common transparent objects. Then SAM could produce high-quality masks with various guidance, including points, boxes, scribbles, and texts. Subsequently, we generate pseudo trimaps with SAM masks and transparency detections. The natural image matting model~\cite{vitmatte} will predict alpha mattes with pseudo trimaps and input RGB images. Besides, since each component of the \themam{} (\thename{}) is accessible, users can easily refine alpha mattes with just a few clicks.

\subsection{Segment with User Interaction}

We utilize the Segment Anything Model (SAM)~\cite{SAM} as the foundation model for segmentation in our approach. SAM is capable of producing high-quality masks through user interactions, including the use of points, bounding boxes, and scribbles. Additionally, we incorporate GroundingDINO~\cite{groundingdino} as our open vocabulary detector (OVD) to enhance image matting using textual guidance. GroundingDINO could provide bounding boxes for SAM based on provided text prompts. Leveraging SAM and GroundingDINO, our proposed method, \thename{}, can generate high-quality masks using more cost-effective guidance methods compared to the traditional trimap.

\begin{equation}
    m=
    \begin{cases}
        SAM(i, g) & if~g\in [ P, B, S ] \\
        SAM(i, OVD(g)) & if~g\in [ T ]
    \end{cases}
    \label{eq:interactive_seg}
\end{equation}, while \eq{P, B, S, T} denote points, boxes, scribbles and texts respectively. 

Specifically, given an input RGB image \eq{i} and user interaction guidance \eq{g}, \thename{} will get segmentation mask \eq{m} with pre-trained SAM and OVD. As shown in Equation~\ref{eq:interactive_seg}, when \eq{G} belongs to points, boxes, or scribbles, we can generate the corresponding mask \eq{m} directly from SAM. Since SAM cannot process text information directly, when \eq{g} is text, we use the OVD model to generate the corresponding bounding box. Then we use the generated box to guide SAM to generate the corresponding mask \eq{m}.



\subsection{Pseudo Trimap Generation Algorithm}
\label{method:trimap_gen}

Given the mask \eq{m} generated by SAM, we design a pseudo-trimap generation algorithm.
The objective is to generate pseudo trimaps for natural image matting models~\cite{vitmatte} that closely resemble manually drawn trimaps.

\noindent
\textbf{Basic Trimap.}
Our trimap generation could be divided into two main steps. In the first step, we generate a basic trimap with SAM mask. We treat each object to be opaque in this part and just simply erode and dilate its mask. Specifically, given a binary input mask $m$. We first erode and dilate it to $m_e$ and $m_d$. Then we generate a basic trimap $t_b$ with the Equation~\ref{eq:trimap_basic}:

\begin{equation}
    t_b{(x,y)}=
    \begin{cases}
    1 & if~(x, y) \in m_e \\
    0.5 & if~(x, y) \in m_d, (x, y) \notin m_e \\
    0 & if~(x, y) \notin m_d, (x, y) \notin m_e
    \end{cases}
    \label{eq:trimap_basic}
\end{equation}

\noindent
\textbf{Transparency Correction.}
However, in the case of transparent objects, the transparent region does not necessarily exist only in the edge region of the object.
It means when using Equation~\ref{eq:trimap_basic} for opaque objects, the pseudo trimap may \emph{conflict with Prior 2} mentioned in Section~\ref{sec:prior_trimap}. To address this issue, in the second step, we propose a transparency correction method for the basic trimap, resulting in a corrected trimap denoted as $t_c$. We provide two options for the user. The first option is fully automatic: we utilize a Large Language Model (LLM)~\cite{gpt-4} to generate a list of common transparent objects. Subsequently, we employ an Open Vocabulary (OV) detector~\cite{groundingdino} to identify transparent objects based on the generated list. If the detection results are empty, we consider $t_b$ as the final trimap, i.e., $t_c = t_b$. However, if we detect transparent objects and obtain their corresponding bounding boxes $B_t$, the basic trimap will be refined using Equation~\eqref{eq:trimap_corrected}.

\begin{equation}
    t_c(x, y) = 
    \begin{cases}
        t_b(x, y) & if~(x, y) \notin B_t \\
        0.5 & if~(x, y) \in B_t
    \end{cases}
    \label{eq:trimap_corrected}
\end{equation}
Incorrect transparency predictions can significantly affect the final performance. When there is an error in the OV detector, we give another very simple solution to refine. Users can determine whether an object is transparent or not with just one click. If it is transparent, we generate a corrected pseudo trimap with Equation~\eqref{eq:trimap_user_corrected}.

\begin{equation}
    t_c(x, y) = 
    \begin{cases}
        0.5 & if~t_b(x,y)>0 \\
        0 & if~t_b(x, y)=0
    \end{cases}
    \label{eq:trimap_user_corrected}
\end{equation}

\subsection{Natural Image Matting}
Through the aforementioned steps, we obtain an automatically generated high-quality trimap, which is a key component for high-performance nature image matting. To achieve superior image matting performance, we adopted the state-of-the-art nature image matting method, ViTMatte~\cite{vitmatte}. Specifically, we use state-of-the-art natural image matting methods without any additional finetuning. As shown in Equation~\ref{eq:matting_model}, the pseudo trimap \eq{t_c} and the RGB image \eq{i} will be sent to the matting model. Then we get the final alpha matte prediction \eq{pred_\alpha}. We observe that with proper trimap, they could achieve remarkable results in the wild to matte anything.

\begin{equation}
    pred_\alpha = matting\_model(i, t_c)
    \label{eq:matting_model}
\end{equation}

\section{Experiments}

\subsection{Implementation Details}

The \thename{} comprises three pre-trained models: the interactive segmentation model SAM~\cite{SAM}, the open-vocabulary detection model GroundingDINO~\cite{groundingdino}, and the natural image matting model ViTMatte~\cite{vitmatte}. In terms of model sizes, we employ the ViT-H SAM model, GroundingDINO-T, and ViTMatte-B. When generating the initial trimap using erosion and dilation, we set the kernel size to 15 and allow users to fine-tune it within the range of [1, 100]. The number of iterations of erosion and dilation is fixed at 5. For transparency prediction, we define an initial vocabulary of transparent objects and require GroundingDINO to correct basic trimap with Equation~\ref{eq:trimap_corrected}. Users are also granted the authority to rectify any errors made by GroundingDINO through a single click. All of our experiments have been conducted in a single RTX 3090 GPU. The majority of our interactive processes can be completed with just a few mouse clicks, eliminating the need for complex tasks such as trimap drawing.

\vspace{-3mm}
\subsection{Datasets}

The goal of Matte Anything is to provide high-quality matting results based on users' simple instructions in any given scenario.
We assess the performance of the \themam{} (\thename{}) by conducting evaluations on four distinct datasets that span various dimensions. These datasets include synthetic natural image data, real natural image data, real human image data, and real animal image data. 

\paragraph{Composition-1k}~\cite{DIM} is an extensively utilized benchmark in the field of natural image matting. It consists of a synthetic dataset comprising 50 distinct foreground objects combined with 1000 diverse backgrounds sourced from the COCO~\cite{mscoco} dataset. This dataset serves as an evaluation platform for assessing the performance of \thename{} on synthetic datasets. It enables direct comparisons with previous matting methods based on trimaps and approaches relying on alternative guidance.

\paragraph{AIM-500}~\cite{aim-500} is a dataset comprising real natural images. It consists of 500 high-quality natural images along with their corresponding foreground objects and alpha mattes. The foreground objects in this dataset encompass both opaque entities such as humans and animals, as well as transparent objects like glass cups. AIM-500 serves as a valuable resource for evaluating algorithmic performance in terms of generalization on real images, as well as assessing the capability to handle foreground objects with diverse attributes.

\paragraph{P3M-500~\cite{P3M} and AM-2k~\cite{GFM}}
These two datasets are task-specific image matting datasets, where the foreground is limited to a specific category. For example, the P3M dataset focuses solely on human matting, while the AM-2k dataset concentrates on animal matting. Both types of datasets provide two versions of test sets, namely synthetic image test sets and real image test sets. In our approach, we primarily utilize real images to evaluate the zero-shot performance of our method.

\subsection{Comparison with Previous Image Matting}

\begin{figure}[t]
    \centering
    \includegraphics[width=\textwidth]{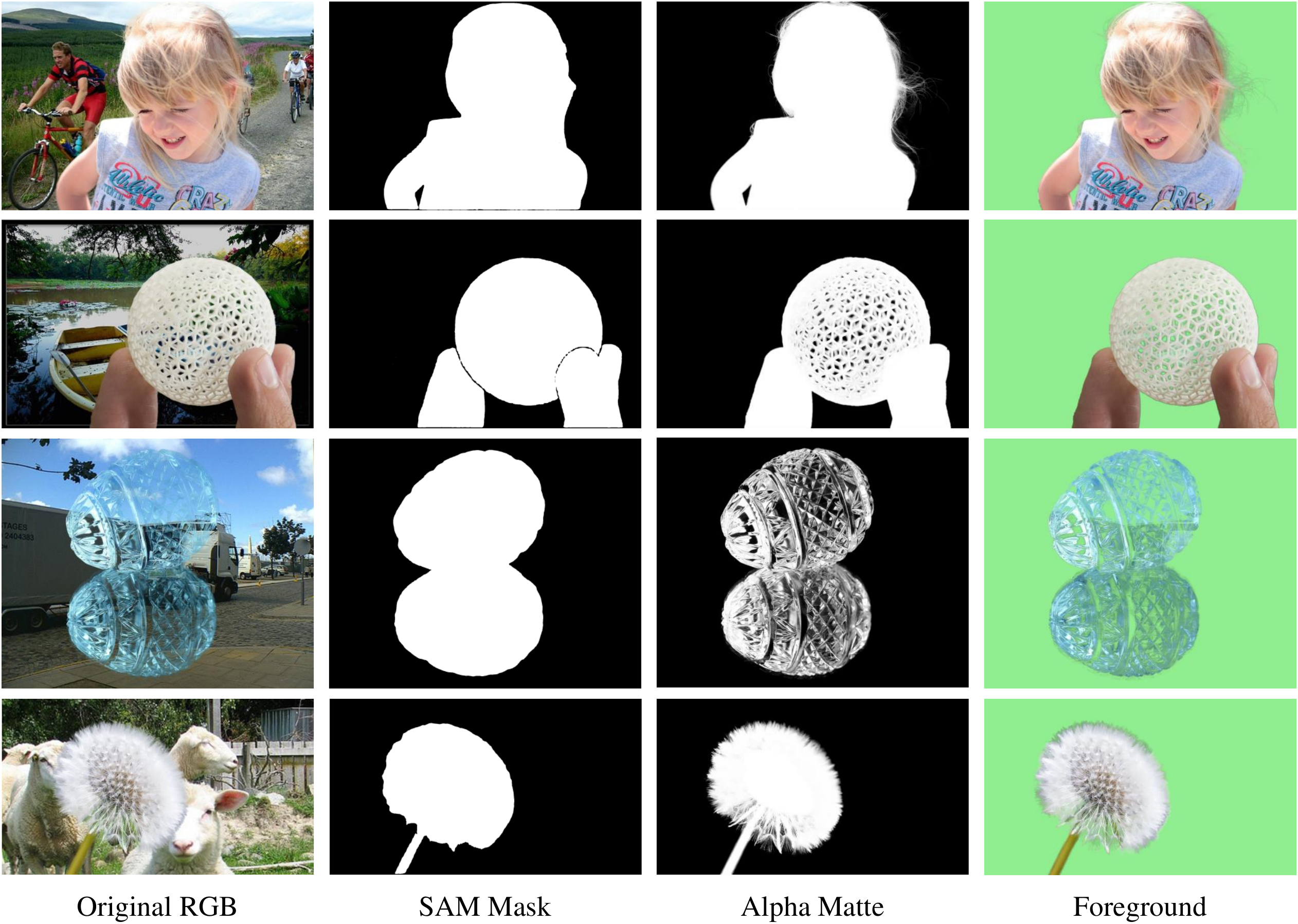}
    \caption{\textbf{Performance on Synthetic Images.}}
    \label{fig:infer_composition-1k}
\end{figure}

To address the scarcity of training data for image matting, a series of deep learning matting methods, exemplified by DIM~\cite{DIM}, have adopted randomly synthesized datasets for training and testing purposes. The synthetic dataset Composition-1k~\cite{DIM} has emerged as the most widely utilized benchmark for evaluating image matting techniques. We evaluate our \thename{} on Composition-1k with various matting methods, as shown in Table~\ref{tab:composition-1k}. Figure~\ref{fig:infer_composition-1k} shows our visualization results. Our experiments reflect the following facts.

\begin{table}[tbp]
    \centering
    \renewcommand{\arraystretch}{1.1}
    \scalebox{0.75}{
    \begin{tabular}{lc|ccc|ccc}
    \toprule
    \multirow{2}{*}{\textbf{Methods}} & \multirow{2}{*}{\textbf{Guidance}} & \multicolumn{3}{c|}{\textbf{SAD}$\downarrow$} & \multicolumn{3}{c}{\textbf{MSE}$\downarrow$} \\
    & & \centering\textit{all} & \textit{transparent} & \textit{opaque} & \centering\textit{all} & \textit{transparent} & \textit{opaque} \\
    \midrule
    \multicolumn{8}{c}{\textbf{Trimap Based Methods}} \\
    \midrule
    DIM~\cite{DIM} & Tri & \centering59.6 & 122.5 & 14.0 & \centering8.5 & 17.9 & 1.7 \\
    IndexNet~\cite{indexnet} & Tri & \centering45.7 & 92.9 & 11.6 & \centering5.2 & 10.9 & 1.1 \\
    MatteFormer~\cite{matteformer} & Tri &  \centering23.8 & 46.7 & 7.2 & \centering1.3 & 2.6 & 0.4 \\
    ViTMatte~\cite{vitmatte} & Tri & \centering20.4 & 39.8 & 6.3 & \centering1.1 & 1.9 & 0.5 \\
    \midrule
    \multicolumn{8}{c}{\textbf{Trimap Free Methods}} \\
    \midrule
    LFMatting~\cite{LFM} & - & \centering58.3 & - & - & \centering11.0 & - & - \\
    HAttMatting~\cite{HAttMatting} & - & \centering44.1 & - & - & \centering7.0 & - & - \\
    UIMatting~\cite{unimatting} & P,B,S & \centering49.9 & - & - & \centering6.0 & - & - \\
    ClickMatting~\cite{user_click} & P & \textcolor{my_purple}{16.8} & - & - & \textcolor{my_purple}{3.1} & - & - \\
    \rowcolor{matany!20}
    \textbf{\thename{}(Ours)} & ~P,B,S,T~ & 39.7 & 78.3 & 11.8 & 6.6 & 13.9 & 1.3 \\
    \rowcolor{matany!20}
    \textbf{\thename{}(Ours)\dag} & P,B,S,T~& \textbf{26.2} & \textbf{50.1} & \textbf{8.8} & \textbf{2.5} & \textbf{5.2} & \textbf{0.5} \\
    \rowcolor{matany!20}
    \textbf{\thename{}(Ours)\dag} & P,B,S,T~& \textcolor{my_purple}{\textbf{6.1}} & \textcolor{my_purple}{\textbf{11.4}}  & \textcolor{my_purple}{\textbf{2.3}} & \textcolor{my_purple}{\textbf{2.3}} & \textcolor{my_purple}{\textbf{4.4}} & \textcolor{my_purple}{\textbf{0.7}} \\
    \bottomrule
    \end{tabular}}
    \caption{\textbf{Results on the Composition-1k.~\cite{DIM}} \textcolor{matany}{\thename{}} is the top-performing trimap-free method, achieving a new state-of-the-art (SOTA). Simultaneously, it has demonstrated highly competitive performance comparable to trimap-based methods. Tri, P, B, S, and T respectively represent trimap, point, box, scribble, and text. \dag~represents results with transparent correction. Results in \textcolor{my_purple}{purple} are testes with smaller image size, staying consistent with~\cite{user_click}}
    \label{tab:composition-1k}
\end{table}

\begin{table}[]
    \centering
    \renewcommand{\arraystretch}{1.1}
    \scalebox{0.7}{
    \begin{tabular}{lc|m{3em}cc|m{3em}cc}
    \toprule
    \multirow{2}{*}{\textbf{Methods}} & \multirow{2}{*}{\textbf{Guidance}} & \multicolumn{3}{c|}{\textbf{SAD}$\downarrow$} & \multicolumn{3}{c}{\textbf{MSE}$\downarrow$} \\
    & & \centering\textit{all} & \textit{transparent} & \textit{opaque} & \centering\textit{all} & \textit{transparent} & \textit{opaque} \\
    \midrule
    GCAMatting~\cite{GCAMatting} & Tri & \centering34.9 & 198.7 & 13.4 & \centering12.1 & 74.9 & 3.8 \\
    IndexnetMatting~\cite{indexnet} & Tri & \centering28.4 & 150.2 & 12.4 & \centering8.6 & 50.9 & 3.0 \\
    MatteFormer~\cite{matteformer} & Tri & \centering26.9 & 147.0 & 11.1 & \centering8.7 & 54.5 & 2.7 \\    
    ViTMatte~\cite{vitmatte} & Tri & \centering17.2 & 80.9 & 8.9 & \centering3.8 & 21.3 & 1.5 \\
    \rowcolor{matany!20}
    \textbf{\thename{}(Ours)\dag} & \textbf{P,B,S,T} & \centering27.8 & 110.5 & 17.0 & \centering9.4 & 35.3 & 6.0 \\
    \bottomrule
    \end{tabular}
    }
    \caption{\textbf{Zero-shot performance on AIM-500.~\cite{aim-500}} \textcolor{matany}{\thename{}} achieves comparable performance against previous trimap-based methods with much simpler user interaction. Tri, P, B, S, and T respectively represent trimap, point, box, scribble, and text.}
    \label{tab:aim-500}
\end{table}

\textbf{SOTA Performance in Trimap-Free Methods.}
Firstly, \thename{} outperforms all existing \textit{trimap-free} methods, establishing a new state-of-the-art (SOTA) results. Specifically, our approach attains remarkable results with a SAD of 26.2 and an MSE of 2.5, surpassing the performance of prior best methods~\cite{unimatting} by 40.6\% and 58.3\%, respectively. The \textcolor{my_purple}{purple} text indicates results obtained after resizing the image. Our method significantly outperforms~\cite{user_click}, reaching 6.1 SAD and 2.3 MSE, with corresponding performance improvements of 63.7\% and 25.8\%. Besides, \thename{} is the trimap-free method with the highest support for interactive means. Boosted by SAM~\cite{SAM} and Grounding DINO~\cite{groundingdino}, our method could be prompted by points, boxes, scribbles, and texts.

\textbf{Very Comparable to Trimap-Based Methods.}
Furthermore, \thename{} achieves very comparable results when compared to \textit{trimap-based} methods. It outperforms DIM~\cite{DIM}, IndexNet~\cite{indexnet}, and GCAMatting~\cite{indexnet}. Compared to the previous state-of-the-art method ViTMatte~\cite{vitmatte}, \thename{} exhibits only a 1.4 decrease in MSE, demonstrating the effectiveness of our pseudo-trimap generation strategy.

\textbf{Simple and Effective in Refinement.}
\thename{} is an interactive matting method, its performance could be significantly refined through simple user interactions. In Table~\ref{tab:composition-1k}, \textit{\thename{}} and \textit{\thename{}\dag} respectively denote the first results and the results after a secondary refinement. Specifically, we first obtain a high-quality mask through user interaction and generate a pseudo-trimap based on Equation~\ref{eq:trimap_basic} and \ref{eq:trimap_corrected}, resulting in outcome \textit{\thename{}}. However, we observe that GroundingDINO achieves a transparency prediction accuracy of 79.9\% on the Composition-1k dataset. Incorrect predictions of transparency can lead to substantial performance degradation. In \thename{}, users can correct these erroneous predictions with just a single click. By applying this simple correction with Equation~\ref{eq:trimap_user_corrected}, we obtain the results of \textit{\thename{}\dag}. It is evident that \textit{\thename{}\dag} exhibits a significant improvement over \textit{\thename{}}, with a SAD improvement of over 34.0\% and an MSE improvement of over 62.1\% for transparent objects.

We visualize the matting results of \thename{} on Composition-1k as illustrated in Figure~\ref{fig:infer_composition-1k}. It can be observed that, in comparison to the segmentation results generated by SAM, our approach produces more intricate and visually satisfying visual effects. Particularly, our method demonstrates significant advantages when dealing with complex and transparent objects. These results demonstrate the superiority of \thename{} in generating refined results, and the effectiveness in dealing with transparent objects.

\subsection{Zero-Shot on Real-World Images}

To evaluate the generalization capability of our model on real-world natural images, we conduct experiments on the AIM-500 dataset using our proposed \thename{} approach. Due to the unavailability of code for some methods mentioned in Table~\ref{tab:composition-1k}, we focused our comparison exclusively on high-performance trimap-based matting methods. Furthermore, as previously discussed, we apply a straightforward refinement process to our results.

\begin{figure*}[t]
    \includegraphics[width=\linewidth]{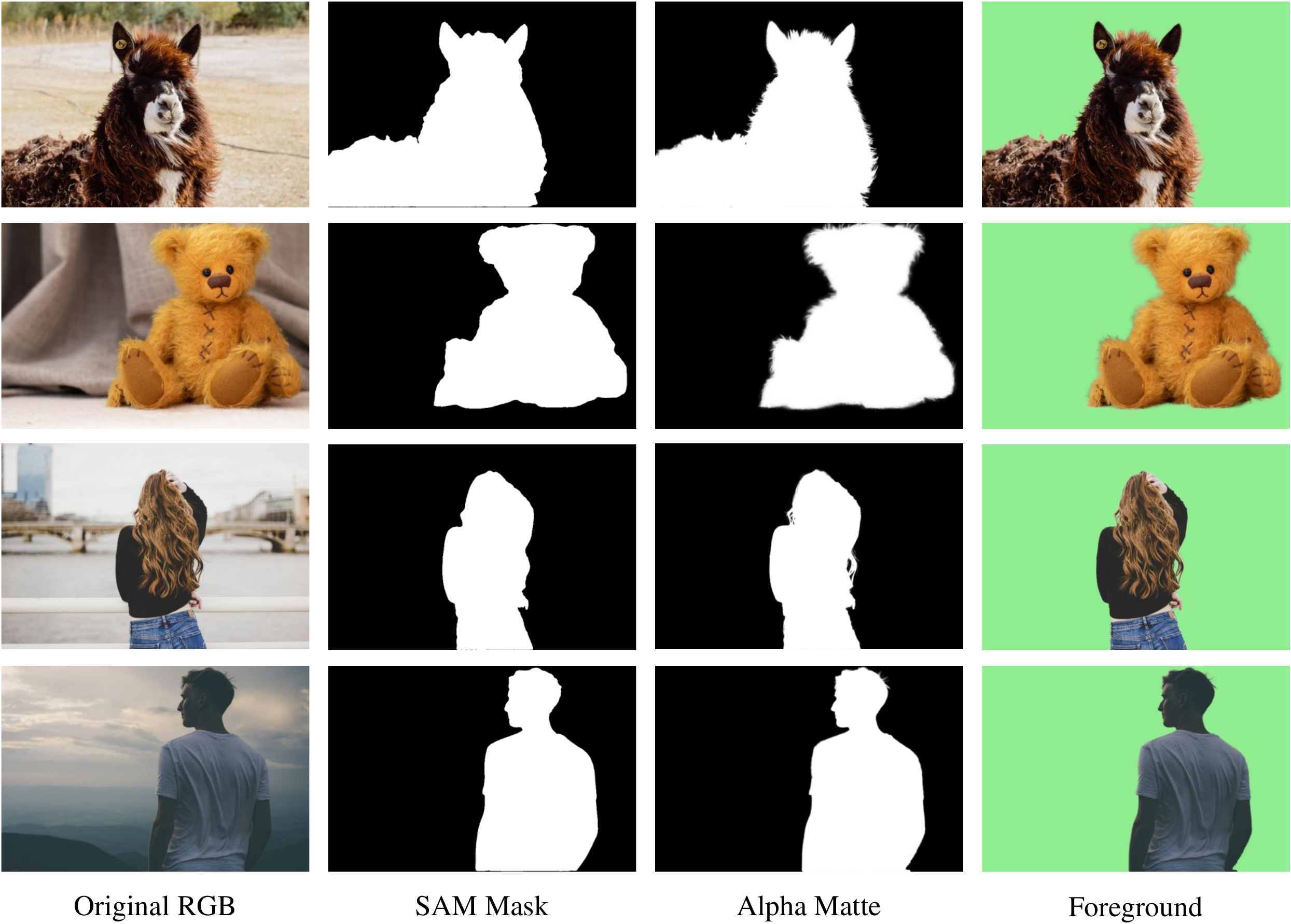}
    \caption{\textbf{Zero-Shot Performance on Real World Images.}}
    \label{fig:infer_aim500}
\end{figure*}

The effectiveness of our approach on the AIM-500 dataset is demonstrated in Table~\ref{tab:aim-500}. Our method showcases promising performance, surpassing certain previous trimap-based matting methods~\cite{GCAMatting, indexnet} and achieving competitive results with MatteFormer~\cite{matteformer} (only 0.9 lower on SAD and 0.7 lower on MSE). As shown in Figure~\ref{fig:infer_aim500}, we visualize the matting results of \thename{} on real-world images.  It could be seen that our model has yielded visually satisfying outcomes when applied to real-world datasets. In contrast to the original Segment Anything model, the results produced by \thename{} exhibit a more refined handling of image details. These show the strong generalization ability of our method.

\subsection{Zero-Shot on Task-Specific Image Matting}

Image matting problems can be further divided based on different matting objectives, such as human image matting, animal image matting, etc. \thename{} is a universal image matting method that can be employed to handle all types of matting problems. Evaluating the performance of \thename{} on these specific category matting tasks is also a focal point of our attention.

We use the test sets of AM-2k and P3M to evaluate the performance of \thename{}. During testing, \thename{} is not subjected to any fine-tuning on the respective datasets, and zero-shot results are reported. As shown in Table~\ref{tab:zs-ts-matting}, \thename{} surpasses most of the task-specific models and achieves very comparable performance against previous sota methods. Specifically, in animal image matting, \thename{} ranks second in quantitative results on image matting, outperforming LFM~\cite{LFM} and SHM~\cite{SHM} (8.1 improvements on SAD and 4.4 improvements on MSE metrics) and slightly underperforming to \cite{GFM}(only 1.6 lower on SAD and 0.9 lower on MSE metrics). On human matting, notably, \thename{} achieves the best quantitative result on the Connectivity metric, 7.1 higher than the previous best method~\cite{GFM}. Besides, \thename{} also achieves very comparable performance to the previous sota method \cite{P3M}.

Overall, even though \thename{} is not trained on the datasets specific to these tasks, it achieves very competitive results compared to previous task-specific matting methods, demonstrating powerful zero-shot performance. This attests to the significant potential of our approach.

\begin{table}[tbp]
    \centering
    \renewcommand{\arraystretch}{1.1}
    \scalebox{0.8}{
    \begin{tabular}{lcc|cccc}
    \toprule
         \textbf{Methods} & \makecell[c]{\textbf{Training Data}} & ~\textbf{Guidance} & \textbf{SAD}$\downarrow$ & ~\textbf{MSE}$\downarrow$ & ~\textbf{Conn}$\downarrow$ & ~\textbf{Grad}$\downarrow$ \\
    \midrule
    \multicolumn{7}{c}{\textbf{Animal Image Matting}} \\
    \midrule
        LFM~\cite{LFM} & ~AM-2k~ & - & 36.1 & 11.6 & 21.1 & 33.6 \\
        SHM~\cite{SHM} & ~AM-2k~ & - & 17.8 & 6.8 & 12.5 & 17.0 \\ 
        GFM~\cite{GFM} & ~AM-2k~ & - & \textbf{9.7} & \textbf{2.4} & \textbf{9.4} & \textbf{9.0} \\ 
        \rowcolor{matany!20}
        {\textbf{\thename{}(Ours)}} & - & P,B,S,T & \underline{11.9} & \underline{3.3} & \underline{10.9} & \underline{11.7} \\ 
    \midrule
    \multicolumn{7}{c}{\textbf{Humam Matting}} \\
    \midrule
        LFM~\cite{LFM} & ~P3M~ & - & 32.6 & 13.1 & 19.5 & 31.9 \\ 
        SHM~\cite{SHM} & ~P3M~ & - & 20.8 & 9.3 & \underline{17.1} & 20.3 \\ 
        GFM~\cite{GFM} & ~P3M~ & - & 15.5 & 5.6 & 18.0 & \textbf{14.8} \\ 
        P3M~\cite{P3M} & ~P3M~ & - & \textbf{9.1}  & \textbf{2.8}  & - & - \\
    \rowcolor{matany!20}
        \textbf{\thename{}(Ours)} & - & P,B,S,T & \underline{10.7} & \underline{2.8} & \textbf{10.0} & \underline{}{17.3} \\ 
    \bottomrule
    \end{tabular}
    }
    \caption{\textbf{Zero-shot performance on task specific image matting.} \textcolor{matany}{\thename{}} could be transferred to animal and human image matting without any additional fine-tuning. It achieves comparable performance to task-specific models and has rich interaction ways.}
    \label{tab:zs-ts-matting}
\end{table}

\subsection{Performance on Multiple Instances Matting}

The non-guided methods~\cite{GFM, P3M, LFM, SHM} mentioned in the previous section exhibit a notable limitation in their paradigm. That is: they treat the matting problem as an image-level task rather than an instance-level task. All the images used in their training and testing datasets feature only a single object (either an animal or a human). This limitation renders these methods unable to handle scenarios involving multiple instances.
Differently, \thename{} is an interactive matting system that emphasizes the importance of user interaction in achieving the desired alpha matte, especially when dealing with multiple objects.

\begin{figure}[t]
    \centering
    \includegraphics[width=\textwidth]{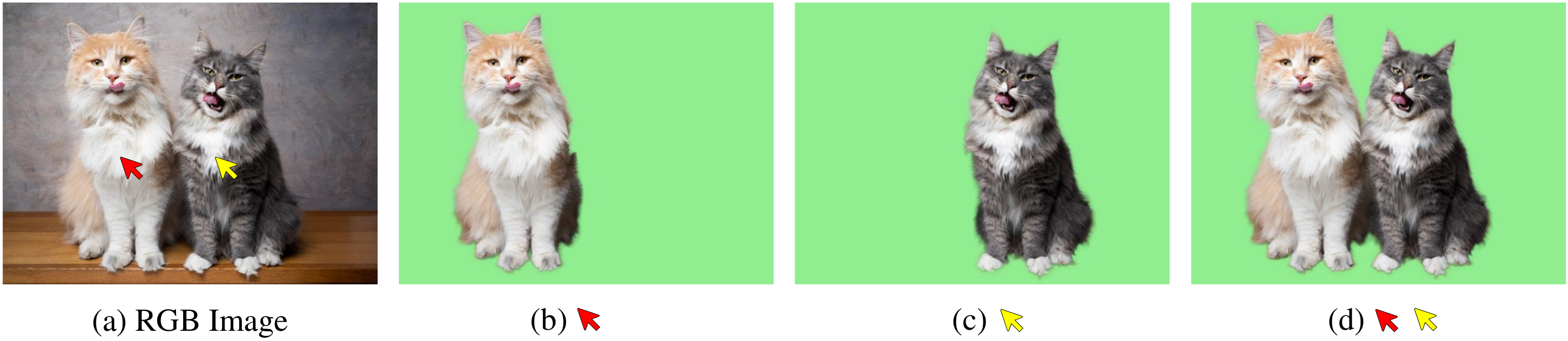}
    \caption{\textbf{Matte with user interaction.} \thename{} could handle the cases with multiple instances.}
    \label{fig:demo5}
\end{figure}

As illustrated in Figure~\ref{fig:demo5}, the resulting mattes vary based on different user interactions. For example, if the user clicks on only one cat on the left or right, \thename{} will matte only that specific cat as instructed. Conversely, if the user clicks on both cats, \thename{} will matte both of them accordingly. This unique advantage of \thename{} showcases its ability to matte any desired instance through simple user clicks, which is challenging to achieve using traditional trimap-based methods.

\section{Ablations}

\subsection{Accuracy of Open-Vocabulary Detector}

\begin{table}[t]
    \centering
    \scalebox{0.9}{
    \begin{tabular}{lcc|c}
    \toprule
        \makecell[l]{\textbf{Transparenct} \\\textbf{Vocabulary}} & \makecell[c]{\textbf{Box Thrd}}~&~\textbf{Backbone}~~&~\textbf{Accuracy} \\
    \midrule
        \text{['glass']} & 0.5 & \multirow{5}{*}{Swin-T~\cite{swin}} & 73.3\% \\ 
        \text{['glass', 'web']} & 0.5 &  & 69.1\% \\ 
        \text{['glass', 'web', 'wine']} & 0.5 &  & 72.1\% \\ 
        \text{['glass']} & 0.4 &  & \textbf{79.9\%} \\
        \text{['glass']} & 0.6 &  & 67.8\% \\
    \bottomrule
    \end{tabular}}
    \caption{\textbf{Accuracy of OV detector on Composition-1k.} GroundingDINO~\cite{groundingdino} could achieve near 80\% accuracy in Composition-1k~\cite{DIM}.}
    \label{tab:ov_detector}
\end{table}

In this section, we try to show that using OVD for transparency detection is a feasible option. We observe that different vocabularies given to the OV detector in \thename{} result in different transparency detection abilities. Though we design a refining strategy with Equation~\ref{eq:trimap_user_corrected}, it is still necessary to verify the validity of the OV detector.

As shown in Table~\ref{tab:ov_detector}, we evaluate GroundingDINO with different box thresholds and vocabularies on Composition-1k. Specifically, the performance of GroundingDINO will vary with the vocabulary and hyperparameter settings. Overall, however, its performance is more than satisfactory, achieving an average accuracy of 72.4\% on Composition-1k. Among them, GroundingDINO-T can reach up to 79.9\% accuracy. The results indicate that open-vocabulary detectors could serve as a solution to detect common transparent objects. 

\begin{table}[t]
    \centering
    \begin{tabular}{lc|cccc}
    \toprule
        \textbf{Experiment} & \textbf{Click Times} & \textbf{SAD} & \textbf{MSE} \\
    \midrule
        \textit{baseline} & 1 & 72.3 & 16.1 \\
        \textit{baseline+ovd}& 1 & 39.7 & 6.6 \\
        \rowcolor{matany!20}
        \textit{baseline+user} & 2 & 26.2 & 2.5 \\
    \bottomrule
    \end{tabular}
    \caption{\textbf{Ablation on Transparent Correction.} It can be seen that the transparency correction in \thename{} can effectively improve the final performance of matting. \textit{baseline} represents experiments  without transparency correction, \textit{baseline+ovd} represents transparency correction with Open-Vocabulary Detector GroundingDINO~\cite{groundingdino}, and \textit{baseline+user} represents the experiments with further user corrections.}
    \label{tab:transparency_correction}
\end{table}

\subsection{Transparent Correction}

\begin{figure}[t]
    \centering
    \includegraphics[width=0.9\linewidth]{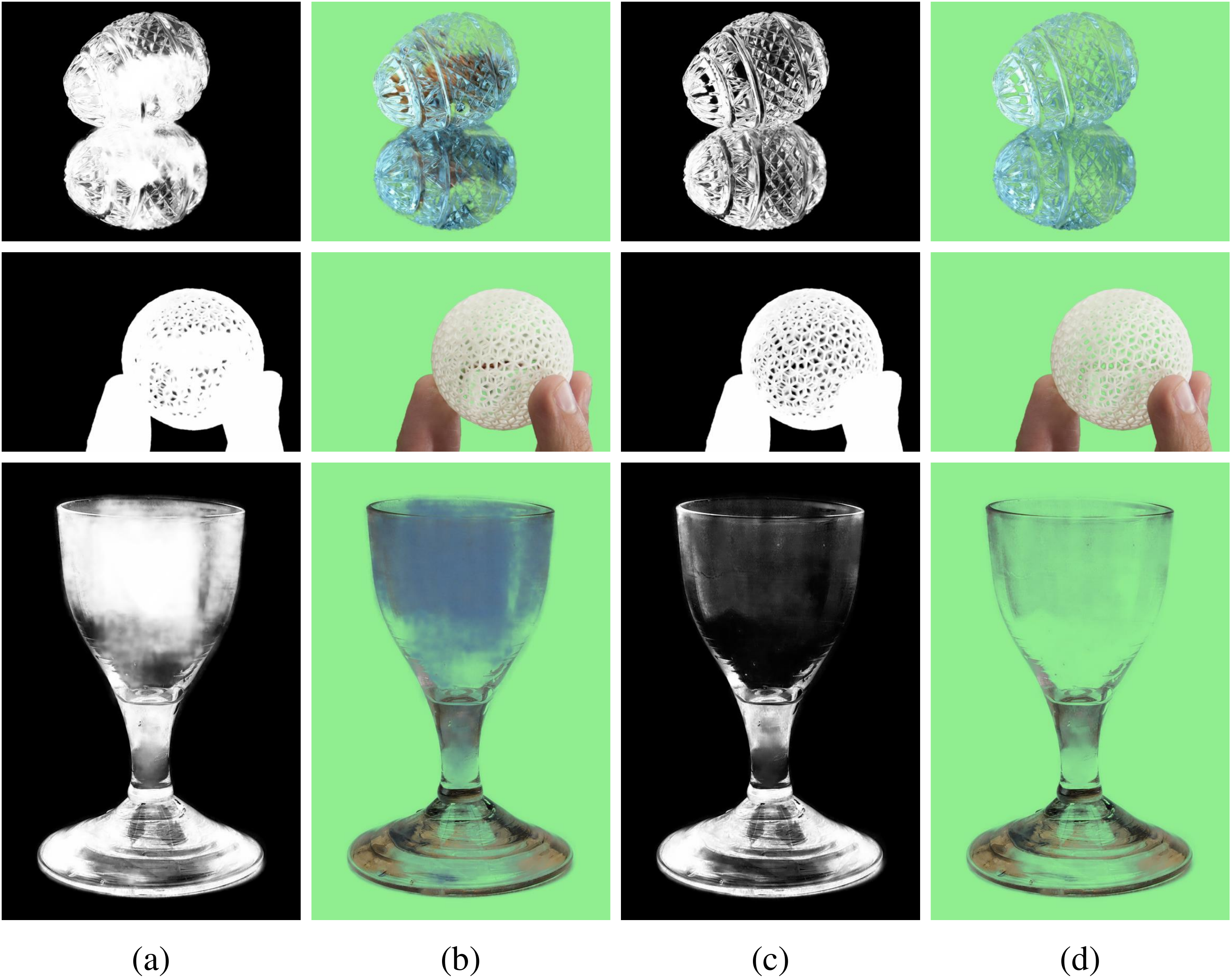}
    \caption{\textbf{Necessity of Transparency correction.} (a) and (b) indicate the alpha mattes and foregrounds without transparency correction, while (c) and (d) indicate the alpha mattes and foregrounds with transparency correction.}
    \label{fig:matany_abs_fig}
\end{figure}

Moreover, we proceed to systematically examine the influence of transparency correction discussed in Section~\ref{method:trimap_gen} on the performance of \thename{}.
The treatment of transparent objects poses an enduring challenge in the domain of image matting~\cite{transmatting}, necessitating a model with a comprehensive grasp of image semantics. In conventional matting methods, the transparent parts of an image are typically confined to the `\textit{unknown region}' delineated in the trimap. In our approach, to generate a pseudo-trimap that satisfies this property, we propose a transparency correction. Here, we provide both quantitative and qualitative explanations of it.

As shown in Figure~\ref{fig:matany_abs_fig}, we compare the impact of transparency correction on the image. In Figures (a) and (b), the alpha matte and foreground are depicted without transparency correction, while Figures (c) and (d) showcase the outcomes following transparency correction. In the absence of transparency correction, the trimap value in the central region of the image is set to 1. This causes the matting model to incorrectly treat them as opaque regions, leading to inaccurate predictions. Through the application of transparency correction, the network is able to rectify this discrepancy, yielding results that are both precise and aesthetically pleasing.

As discussed in Section~\ref{method:trimap_gen}, the transparency correction could be divided into two parts. First, we use OVD to automatically correct the trimap. Then, the user could correct the wrong prediction of OVD with a single click.
As illustrated in Table~\ref{tab:transparency_correction}, when we apply OVD for automatic correction of trimap, the performance of \thename{} improves from 72.3 to 39.7 in terms of SAD. Simultaneously, correcting erroneous transparency predictions with a single click enhanced the network's performance further to 26.2 on SAD, reaching a total improvement of 63.8\%. This result significantly surpasses the previously best-performing click-interaction method, providing evidence of the superior performance of \thename{} and the validity of transparency correction.

\section{Limitations and Future Work} 
The major limitation of Matte Anything lies in the substantial computational burden introduced by employing Segment Anything Models. Since segmentation models primarily offer rough segmentation masks that can be refined by users, the quality of the matting results largely depends on the performance of the image matting model. Therefore, we believe that SAM-H is likely to be replaced by smaller SAM-like models, without significantly compromising the performance of MatAnything. We leave this as our future work.

\section{Conclusion}
In this paper, we present \themam{} (\thename{}), a highly performant interactive matting system designed to address the labor-intensive process of trimap generation. The key insight of our work is to generate pseudo trimap automatically. Our approach leverages the potential of vision foundation models in the field of image matting. \thename{} comprises three pre-trained models, including the segment anything model, open-vocabulary detector, and natural image matting model, eliminating the need for additional training. By analyzing the properties of the trimap and these underlying models, we can automatically generate a pseudo-trimap that resembles the effect of a real trimap.

We evaluate the performance of \thename{} on four benchmark datasets. Through simple refinement, our method outperforms all existing trimap-free methods and achieves competitive results with trimap-guided methods on the Composition-1k dataset. Moreover, \thename{} demonstrates robust generation capability in various real-world scenarios and zero-shot performance on task-specific image matting scenarios. We hope that our work could inspire future advancements in interactive image matting and find wide application in numerous real-world tasks.

\bibliographystyle{elsart-num}
\bibliography{ref}

\end{sloppypar}
\end{document}